# A Deep-Learning Intelligent System Incorporating Data Augmentation for Short-Term Voltage Stability Assessment of Power Systems

Yang Li [a,1,*], Meng Zhang [b, 1], Chen Chen [c]

[a] School of Electrical Engineering, Northeast Electric Power University, Jilin 132012, China

[b] National Key Laboratory of Science and Technology on Vessel Integrated Power System, Naval University of Engineering, Wuhan 430033, China

[c] School of Electrical Engineering, Xi'an Jiaotong University, Xi'an 710049, China

[*] Corresponding author. E-mail address: liyang@neepu.edu.cn (Y. Li).

[1] Yang Li and Meng Zhang contributed equally to this work and should be considered co-first authors.

**Abstract**—Facing the difficulty of expensive and trivial data collection and annotation, how to make a deep learning-based short-term voltage stability assessment (STVSA) model work well on a small training dataset is a challenging and urgent problem. Although a big enough dataset can be directly generated by contingency simulation, this data generation process is usually cumbersome and inefficient; while data augmentation provides a low-cost and efficient way to artificially inflate the representative and diversified training datasets with label preserving transformations. In this respect, this paper proposes a novel deep-learning intelligent system incorporating data augmentation for STVSA of power systems. First, due to the unavailability of reliable quantitative criteria to judge the stability status for a specific power system, semi-supervised cluster learning is leveraged to obtain labeled samples in an original small dataset. Second, to make deep learning applicable to the small dataset, conditional least squares generative adversarial networks (LSGAN)-based data augmentation is introduced to expand the original dataset via artificially creating additional valid samples. Third, to extract temporal dependencies from the post-disturbance dynamic trajectories of a system, a bi-directional gated recurrent unit with attention mechanism based assessment model is established, which bi-directionally learns the significant time dependencies and automatically allocates attention weights. The test results demonstrate the presented approach manages to achieve better accuracy and a faster response time with original small datasets. Besides classification accuracy, this work employs statistical measures to comprehensively examine the performance of the proposal.

**NOMENCLATURE**
**Acronyms**

| | |
|---|---|
| STVS | Short-term voltage stability |
| STVSA | Short-term voltage stability assessment |
| PMU | Phase measurement unit |
| TS | Time series |
| ANN | Artificial neural network |
| DT | Decision tree |
| ELM | Extreme learning machine |
| RVFL | Random vector functional link network |
| SVM | Support vector machine |
| RF | Random forest |
| RNN | Recurrent neural network |
| CNN | Convolutional neural network |
| GCN | Graph convolutional network |
| GRU | Gated recurrent unit |
| LSTM | Long short term memory |
| BiGRU-Attention | Bi-directional gated recurrent unit with attention mechanism |
| GAN | Generative adversarial network |
| LSGAN | Least squares generative adversarial network |
| CGAN | Conditional generative adversarial network |
| DLBAN | Deep LSGAN-BiGRU-Attention network |
| SFCM | Semi-supervised fuzzy c-means |
| COP-k-means | Constraint-partitioning k-means |
| WD | Wasserstein distance |
| MMD | Maximum mean discrepancy |
| FID | Fréchet inception distance |
| ROC | Receiver operating characteristic |
| AUC | Area under the receiver operating characteristic curve |
| MCC | Matthews correlation coefficient |
| SC | Silhouette coefficient |
| FC | Fully connected |
| TDS | Time-domain simulation |
| OTW | Observation time window |
| PDC | Phasor data concentrator |
| SNR | Signal-to-noise ratio |

**Symbols**

| | |
|---|---|
| $U$ | Membership matrix |
| $V$ | Cluster center |
| $F$ | Supervision information |
| $d$ | Euclidean distance |
| $\alpha$ | Scaling factor |
| $G$ | Generator |
| $D$ | Discriminator |
| $Z$ | Noise signal |
| $y$ | Label information |
| $r_t$ | Reset gate |
| $z_t$ | Update gate |
| $h_t$ | Hidden state |
| $\Delta t$ | Sampling time |
| $\mu$ | Empirical mean |
| cov | Empirical covariance |

**Subscripts**

| | |
|---|---|
| $t$ | Sampling time |
| $N$ | Number of all the instances |
| $L$ | Number of the total buses |
| $T$ | Observation time window size |
| $q$ | Length of time series |
| $c$ | Number of clusters |
| $att$ | Attention mechanism |
| $dense$ | Fully connect layer |

**Superscript**

| | |
|---|---|
| $r$ | Real data |
| $g$ | Generated data |

# I. Introduction

Short-term voltage stability assessment (STVSA) has always been regarded as a critical task to ensure the secure and stable operation of a power system. The short-term voltage stability (STVS) of the power system refers to the ability of the bus voltage to quickly restore to an acceptable level after a large disturbance [1, 2]. From the perspective of pattern recognition, STVSA can be viewed as a binary classification problem [3, 4]. With the reform of electricity markets, decommissioning of aging thermal power plants and rapidly growing load demands, the power transmission capacity is approaching its limit, which seriously threatens the voltage stability of today's power systems [5]. Due to the inherent variability and uncertainty of renewable power sources [6], the increasing penetration of renewable generations and induction motor loads also poses new challenges for the system voltage stability [7]. Therefore, how to detect the STVS status of power systems accurately and timely has become a challenging and urgent problem.

Some pioneering studies have been devoted to solving this STVSA problem, such as stability assessment based on energy function [8] and stability boundary analysis based on PV plane [9]. In reference [10], a model and data hybrid method was proposed for monitoring the STVS status in real time. Although Lyapunov exponents approach can achieve model-free STVSA, the turbulent Lyapunov exponents lingering around 0 will take a long time to obtain a reliable assessment result [11]. Different from the above methods that are heavily reliant on accurate power system physical models, data-driven STVSA approaches based on machine learning have recently attracted growing concerns from academia and industry, as the successful commercial application of wide-area measurement systems makes high-precision synchronized measurements available. Shallow machine learning techniques, such as artificial neural networks (ANNs) [12], decision trees (DTs) [4, 5, 13], extreme learning machines (ELMs) [14], and random vector functional link networks (RVFL) [15, 16] have been reported to deal with the STVSA problem. More recently, state-of-the-art works are able to cope with missing data, knowledge transfer, self-adaptive classification. To be specific, reference [14] realizes a hierarchical self-adaptive STVSA by using an ensemble-based randomized learning model, while reference [16] proposes a missing-data tolerant method for post-fault STVSA. Compared with these shallow networks, emerging deep learning techniques can effectively extract deep features in an end-to-end manner without depending on expert domain knowledge. In the area of security assessment of power systems, some deep learning algorithms have been introduced. In [17], generative adversarial network (GAN) is used to solve the missing-data problem in dynamic security assessment of power system. Deep transfer learning and deep neural network (DNN) are adopted for static security assessment in [18]. Recurrent neural network (RNN) [19] and convolutional neural network (CNN) [20] have been employed for transient stability assessment of power systems. And for the STVSA problem, graph convolutional network (GCN) [3], long short-term memory (LSTM) [21, 22] are also adopted. However, although the deep learning is introduced into the field of STVSA, how to make the deep learning model work well on the small training dataset is still a quite challenging task.

The commonly-used deep learning model RNN is quite efficient in dealing with sequential data mining problems. LSTM and GRU, as 2 variants of RNN, can solve the gradient disappearance and gradient explosion problem suffered by standard RNNs [23]. Moreover, compared with LSTM, GRU has a simpler structure together with fewer parameters, which enables GRU to have higher training efficiency without sacrificing accuracy. Meanwhile, attention mechanisms have proven to be remarkably effective in learning important information about time series (TSs) data. Thus, in this work, bi-directional gated recurrent unit with attention mechanism (BiGRU-Attention) is utilized to capture the time dependencies in post-disturbance system dynamics for STVSA. As is well known, compared with traditional shallow machine learning methodologies, access to a large corpus of training data is a crucial prerequisite to ensure that a deep learning algorithm, as the core of the big data intelligence, can effectively learn complex data distribution characteristics [24, 25]. Due to the lack of a unified and reliable criterion, it is tough and expensive to obtain large-scale, balanced data with accurate labels. Besides data labeling, training data collection itself is expensive and laborious in practical applications, which constitutes an important barrier for developing a deep learning based STVSA model with a high performance in real-world applications. In view of this, it's of great significances to train a deep learning model to enable it to work well on small data in the STVSA field. Unfortunately, to the best knowledge of the authors, there is no research devoted to addressing this problem until now.

Facing the difficulty of the expensive and trivial data collection and annotation, the motivation of this work is to cope with training a deep learning model with small dataset. It's known that data augmentation plays a critical role in training a successful deep learning model since it can increase the diversity of training data. As far as STVSA is concerned, there exist 2 different means to collect large amounts of training data: contingency simulation and data augmentation. Although a big enough data set can be directly generated by contingency simulation, data augmentation is a superior and irreplaceable proposition. In contrast to the cumbersome, complicated and inefficient contingency simulation, data augmentation provides a low-cost and efficient way to artificially inflate the representative and diversified training datasets with label preserving transformations. As an effective extension of original GAN proposed by Goodfellow in 2014 [26], least squares generative adversarial network (LSGAN) was put forward in 2017 [27] and has been successfully for renewable scenario generation due to its powerful generative modelling ability [28]. In view of this situation, this paper puts forward a deep adversarial data augmentation technique based on LSGAN such that the BiGRU-Attention-based STVSA model is able to be applicable to small training datasets.

To highlight the research gaps and contributions of the proposed work in this paper, a comprehensive comparison with the state of the art methods and recent studies has been performed in Table I, where the symbol ✓ and ✗ respectively indicates the relevant references adopt or don't adopt the corresponding method contained in the items.

TABLE I
Comparison of the proposed approach with related works

| References | Items | | | |
| --- | --- | --- | --- | --- |
| | Deep learning | Data augmentation | Labeling data | Statistical measures |
| [3] | ✓ | ✗ | ✗ | ✗ |
| [4] | ✗ | ✗ | ✓ | ✗ |
| [5] | × | × | × | × |
| [14] | × | × | × | × |
| [15] | × | × | × | × |
| [21] | ✓ | × | × | × |
| [22] | ✓ | × | ✓ | ✓ |
| **Proposed approach** | ✓ | ✓ | ✓ | ✓ |

In Table I, the listed items are introduced in detail as follows:

Deep learning: it means that the STVSA model is established based on deep learning techniques.

Data augmentation: it means that the finite small dataset is expanded by using data augmentation.

Labeling data: Before training the STVSA model, the collected data are annotated with the exact class labels information.

Statistical measures: statistical measures, such as operating characteristic curve (AUC), Matthews correlation coefficient (MCC), and F1-score, are used to comprehensively measure the overall characteristics of the constructed assessment model [22, 29, 30].

According to this table and the above literature review, the most important contributions can be highlighted as follows:

(1) The paper presented a novel deep-learning intelligent system incorporating LSGAN-based data augmentation and BiGRU-Attention-based assessment model, called Deep LSGAN-BiGRU-Attention network (DLBAN).

(2) By leveraging data augmentation, the proposed deep learning-based assessment model is able to work well with small training datasets, which is a new methodology in the field of power system stability analysis.

(3) To fully capture the temporal dependence from the input post-disturbance dynamic trajectories of power systems, BiGRU-Attention is utilized to extract the latent information from the forward and backward directions, where attention mechanism improves the feature learning ability through automatic allocation of attention based on the importance of input information.

(4) Due to the unavailability of a reliable quantitative criteria in the STVSA field, semi-supervised fuzzy c-means (SFCM) is adopted to determine the labels of samples whose stability statuses cannot be intuitively distinguished according to domain knowledge.

(5) This work has performed extensive statistical tests using statistical indicators such as AUC, MCC, and F1-score to comprehensively examine the performance of the proposed approach.

## II. Semi-Supervised Cluster Learning

In this study, the purpose of the semi-supervised cluster learning is to label different samples. Although the STVS of power systems can be quantified by many existing indices reflecting the STVS related risk level [2, 12, 14], this work aims to judging whether a power system can hold short-term voltage stability after a large disturbance by using a supervised binary classification model, rather than monitoring the system STVS related risk level. By using labeled samples, the training process of the built supervised deep learning model is realized, and thereby achieving the accurate assessment of the STVS. Unfortunately, there is still no unified reliable criterion to determine the STVS for a given power system up to now, how to efficiently label all training data becomes a bottleneck, which hinders the application of data-driven STVSA methods into real-world power systems [5]. In this context, this section details how to assign a binary label for the samples by using semi-supervised cluster learning.

According to domain knowledge, the labels of some samples can be easily obtained. For example, if all post-contingency bus voltages are above 0.9 or below 0.75, there is no doubt that it can be labeled as stability or instability. Considering that the precisely labeled samples, as prior knowledge, can guide the clustering process to good search spaces, the semi-supervised cluster learning is adopted to obtain all class labels of the dataset, which avoids the blindness of the unsupervised clustering method and a large amount of time required for labelling different samples by using expertise [31].

### A. Semi-Supervised Fuzzy C-Means Algorithm

In this study, SFCM is used to obtain the exact class labels of all the samples [32]. For an $m$-dimensional dataset $X=\{x_j\}, 1 \leq j \leq N$, it is divided into $c$ clusters. And for each sample $x_j=(x_{j1}, x_{j2}, \cdots, x_{jm})$, it is divided into the cluster $S_i$, $1 \leq i \leq c$ by fuzzy membership. In semi-supervised learning, the samples with exact labels are served as the prior knowledge during the iterative optimization process. The objective function of SFCM is given as follows [32]:

$$\min J(U,V) = \sum_{i=1}^{c}\sum_{j=1}^{N} u_{ij}^2 d_{ij}^2 + \alpha \sum_{i=1}^{c}\sum_{j=1}^{N}(u_{ij}-f_{ij}b_j)^2 d_{ij}^2 \quad (1)$$

where $U=[u_{ij}]$ is the membership matrix, $V=(v_1, v_2, \cdots, v_c)$ is the cluster center, where $v_i$ is the $i$th cluster center; $d_{ij}^2$ is the Euclidean distance between $x_j$ and $v_i$; $F=[f_{ij}]_{c\times N}$ represents the supervision information, $f_{ij}$ is the pre-knowledge of $u_{ij}$; the first term of (1) is the objective function of standard FCM and the second term reflects supervised clustering; $\alpha$ ($\alpha \geq 0$) is a scaling factor, which is used to measure the importance of the given classification information; $b=[b_j]^T$ represents whether sample $x_j$ has a known class label, and the condition that $b_j$ needs to meet is

$$\begin{cases} b_j = 1, & x_j \text{ is labled;} \\ b_j = 0, & \text{otherwise.} \end{cases} \quad (2)$$

Specifically, the representation of the membership $u_{ij}$ and the cluster center $v_i$ are as follows:

$$u_{ij} = \frac{1}{1+\alpha}\left[\frac{1+\alpha\left(1-b_j\sum_{i=1}^{c}f_{ij}\right)}{\sum_{j=1}^{c}\frac{d_{ij}^2}{d_{kj}^2}} + \alpha f_{ij}b_j\right] \quad (3)$$

$$v_j = \frac{\sum_{j=1}^{N} u_{ij}^2 x_j}{\sum_{j=1}^{N} u_{ij}^2} \quad (4)$$

The details of the SFCM algorithm used in this work can be found in [32].

### B. Silhouette Coefficient

In this paper, the silhouette coefficient (SC) is adopted to evaluate the performance of the semi-supervised cluster learning algorithm [33]. A larger value of SC implies a better clustering performance, which means that the points of the same cluster are more compact and the points between different clusters are more separated. The SC is defined as [33]:

$$SC = \frac{1}{N}\sum_{j=1}^{N}\frac{b_j - a_j}{\max(a_j, b_j)} \quad (5)$$

where $a_j$ represents the compactness of the inner-cluster and $b_j$ reflects the separation between inter-clusters. For dataset $X=\{x_j\}$, $1 \leq j \leq N$, it is divided into $c$ clusters denoted by $\{S_1, S_2, \cdots, S_c\}$, here $x_j \in S_i$ and $a_j$ represents the average distance between $x_j$ and other points in the inner-cluster; for inter-cluster $S_k(1 \leq k \leq c, k \neq i)$, $D(i,k)$ represents the average distance between $x_j$ and all points in $S_k$, and $b_j = \min\{D(i,k) | 1 \leq k \leq c, k \neq i\}$ is the minimum average distance from $x_j$ to all other clusters.

## III. DLBAN Framework

This paper proposes a novel deep-learning intelligent system combining LSGAN and BiGRU-Attention for STVSA, named DLBAN. The basic principles of LSGAN and BiGRU-Attention are described in detail as follows.

### A. Least Squares Generative Adversarial Network

Compared with the original GAN, the least squares loss function is adopted for discriminator instead of the sigmoid cross-entropy loss function in LSGAN [27]. By doing so, the samples that are far from the decision boundary are penalized and move toward the boundary, thus the vanishing gradients problem is effectively solved. The cost function of LSGAN can be formulated as [28]:

$$\min_D L(D) = \frac{1}{2} E_{x \sim P_{data}(x)}\left[(D(x)-b)^2\right] + \frac{1}{2} E_{Z \sim P_Z(Z)}\left[(D(G(Z))-a)^2\right] \quad (6)$$

$$\min_G L(G) = \frac{1}{2} E_{Z \sim P_Z(Z)} \left[ (D(G(Z)) - c)^2 \right] \quad (7)$$

where $D$ is the discriminator and $P_{data}(x)$ denotes the genuine probability distribution; $G$ is the generator which samples the input noise variable $Z$ from Gaussian distribution $P_Z(Z)$.

In this paper, a conditional version of LSGAN is used where the class labels are binarized by using one-hot encoding [34]. Label $y$ as the condition is attached to noise signal $Z$, which constitutes the generator's input $(Z, y)$. Similarly, label $y$ is appended to real data $x$ as the discriminator's input $(x, y)$. Thereby, the cost functions are rewritten as

$$\min_D L(D) = \frac{1}{2} E_{x, y \sim P_{data}(x,y)} \left[ (D(x, y) - b)^2 \right] + \\ \frac{1}{2} E_{Z \sim P_Z(Z), y \sim P_{data}(x,y)} \left[ (D(G(Z, y), y) - a)^2 \right] \quad (8)$$

$$\min_G L(G) = \frac{1}{2} E_{Z \sim P_Z(Z), y \sim P_{data}(x,y)} \left[ (D(G(Z, y), y) - c)^2 \right] \quad (9)$$

Here, $a$, $b$, and $c$ satisfy the following conditions: $b-c=1$, and $b-a=2$. Specifically, $a$, $b$ and $c$ are taken as 0, 1, and 1.

## B. BiGRU-Attention

BiGRU can fully capture time dependence in forward and backward directions from input information. To improve the performance of the BiGRU, an attention mechanism is adopted to focus on the key hidden information.

### 1) BiGRU

As a variant of RNNs, GRU overcomes the gradient disappearance and gradient explosion problems during training process. Compared with LSTM, GRU is a simpler algorithm with few parameters. The structure of GRU is shown in Fig. 1.

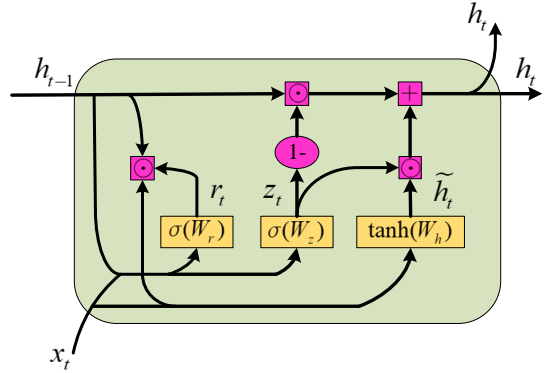

Fig. 1. The structure of GRU

In Fig. 1, $t$ denotes the sampling time at moment $t$; the reset gate $r_t$ determines how much information in the previous state is to be ignored. The update gate $z_t$ decides how much information in the previous state is added to the current state; $r_t$ and $z_t$ are updated as follows [35]:

$$r_t = \sigma(W_r x_t + U_r h_{t-1} + b_r) \quad (10)$$
$$z_t = \sigma(W_z x_t + U_z h_{t-1} + b_r) \quad (11)$$

where $\sigma$ is the sigmoid function; $h_{t-1}$ represents the previous hidden state; at time $t$, the new hidden state is calculated as

$$h_t = (1 - z_t) \odot h_{t-1} + z_t \odot \tilde{h}_t \quad (12)$$

where $\tilde{h}_t$ represents the current candidate state; $\odot$ denotes the element-wise product. The representation $\tilde{h}_t$ is as follows:

$$\tilde{h}_t = \tanh(W_h x_t + r_t \odot (U_h h_{t-1}) + b_h) \quad (13)$$

In equations (10-13), $W$, $U$ and $b$ are the parameters that need to be tuned during the training process of the GRU.

As for the STVSA, both the previous and subsequent states contribute to the output at the current state. Hence, BiGRU is utilized in this study by concatenating forward and backward GRU together to make full use of information in the previous and subsequent states.

2) Attention Mechanism

The attention mechanism allocates enough attention to key information and highlights the impact of important information, thereby the accuracy can be improved [35]. Considering that the system dynamics information at different sampling times contributes to varying degrees for the STVSA, the attention mechanism is adopted to better measure the importance of the extracted hidden layer features and automatically assign corresponding weights to them.

The formulas for calculating the normalized weight of the attention mechanism $s$ can be formulated as follow:

$$u_t = \tanh(W_{att} h_t + b_{att}) \tag{14}$$

$$\alpha_t = \frac{\exp(u_t^{\mathrm{T}} u_{att})}{\sum_t \exp(u_t^{\mathrm{T}} u_{att})} \tag{15}$$

$$s = \sum_t \alpha_t h_t \tag{16}$$

where $u_t$ is a fully connected (FC) layer for learning hidden layer features, followed by a softmax layer which outputs a probability distribution $\alpha_t$; $W_{att}$ and $b_{att}$ denote trainable weights and bias; $u_{att}$ is a randomly initialized context vector; the output of the attention mechanism is represented by $s$. The mechanism selects and extracts the most significant temporal information from hidden layers by multiplying $\alpha_t$ concerning the contribution to the decoding tasks [35].

*C. Structure of DLBAN Framework*

The DLBAN framework proposed in this paper consists of 2 deep learning algorithms LSGAN and BiGRU-Attention. The structure of DLBAN framework is shown as Fig. 2.

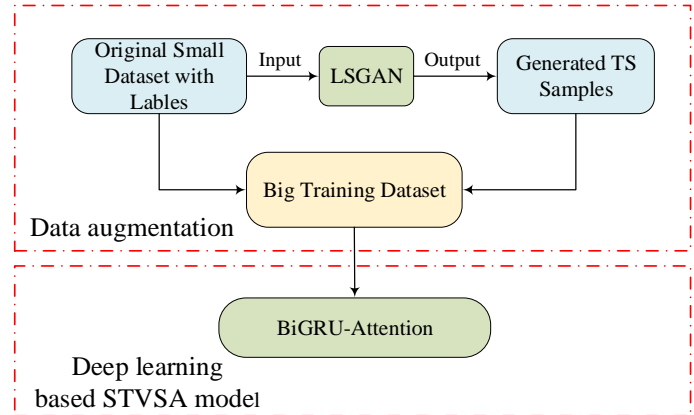

Fig. 2. The structure of DLBAN Framework

In the DLBAN framework, large-scale, reliable, balanced data with accurate labels are generated firstly by LSGAN based on the original small dataset, which is not only increases the amount of training data but also improves the quality of training data. Then, a big training dataset consisting of the generated samples and original small dataset is obtained, achieving a low-cost and efficient data augmentation. Finally, based on the obtained big training dataset, the BiGRU-Attention-based deep learning model that requires massive training data to effectively learn is trained. In this way, the proposed DLBAN framework enables the proposed deep learning-based STVSA model to give full play to its powerful advantages in deep feature mining, even if in case of a small dataset.

## IV. Proposed STVSA Intelligent System

The proposed DLBAN-based STVSA intelligent system is illustrated in Fig. 3.

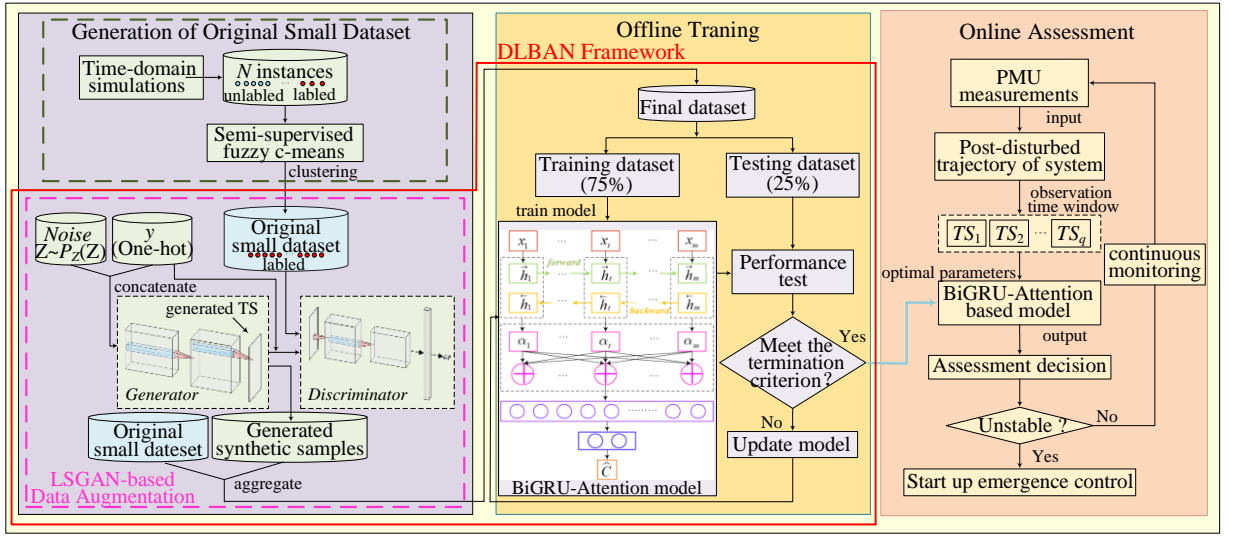

Fig. 3. Flowchart of the proposed STVSA method

From Fig. 3, it can be seen that the implementation of the proposed STVSA method can be divide into 4 stages.

1) Stage 1: the original small dataset is generated by using time-domain simulations (TDSs) and the SFCM;

2) Stage 2: the data augmentation is performed by aggregating the original small dataset and the artificially inflated training set from the LSGAN;

3) Stage 3: the BiGRU-Attention assessment model is built for offline training;

4) Stage 4: the BiGRU-Attention assessment model is utilized for online STVSA.

## A. Generation of Original Small Dataset

In this paper, $N$ instances are generated by TDSs, which can be formulated as

$$D_{TS} = \{TS_1, TS_2, \cdots, TS_j, \cdots TS_N\}, j \in [1, N] \tag{17}$$

In this formula, $N$ is the number of all the instances.

According to the domain knowledge, the voltage trajectories shown in Fig. 4 can be straightforwardly classified as stable or unstable. Here, during the process of semi-supervised clustering learning based on the SFCM, a part of samples with known labels obtained according to domain knowledge are served as prior information for data annotation.

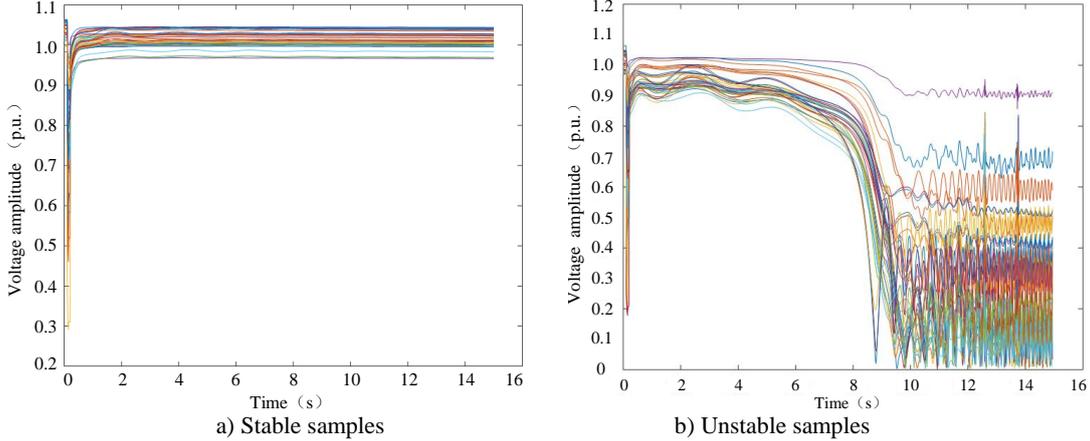

a) Stable samples　　　　b) Unstable samples

Fig. 4. Examples of different STVS samples

And then, an original small dataset is obtained, in which all samples have class labels. In the original small dataset, the input TSs of $N$ instances are composed of the 3 electrical quantities which are strongly related to the STVS status. In this paper, the 3 quantities are respectively the bus voltage amplitude, the active power, and the reactive power (abbreviated as U/P/Q), denoted by:

$$TS_j = \begin{cases} \{TS_{j,1},\cdots,TS_{s,L}\}, & \text{for } U \\ \{TS_{j,L+1},\cdots,TS_{j,2L}\}, & \text{for } P \\ \{TS_{j,2L+1},\cdots,TS_{j,3L}\}, & \text{for } Q \end{cases} \quad (18)$$

$$TS_j = \{U_{j,1},U_{j,2},\cdots,U_{j,t},\cdots,U_{j,q},P_{j,1},P_{j,2},\cdots,P_{j,t},\cdots,P_{j,q},Q_{j,1},Q_{j,2},\cdots,Q_{j,t},\cdots,Q_{j,q}\}, 1<t<q \quad (19)$$

where $TS_j$ is the TSs collection of the $j$ th sample of $D_{TS}$; $L$ is the number of the total buses; the dimension of $TS_j$ is $3L$; here, the length of each TS is the same. In (19), $q=T/\Delta t$ is the TS length, $T$ is the size of an observation time window (OTW) and $\Delta t$ denotes sampling time. In this study, the length of OTW is selected as 0.03s through the sensitivity analysis, which will be presented in detail in the follow-up case studies.

*B. Data Augmentation*

In this work, the LSGAN is utilized for data augmentation, which plays an important role as a link between small dataset and deep learning techniques in the proposed intelligent system. To be specific, data augmentation uses the original small data set obtained by TDSs and the SFCM as its input, and outputs massive labeled data with high quality, which provides a necessary prerequisite in term of data resources for training deep learning classification models. The big training dataset obtained by data augmentation serves as the input of the BiGRU-Attention based assessment model. By this means, the data augmentation step supports the STVS stability assessment in this study.

The detailed architecture of the LSGAN used in this work is shown in Fig. 5.

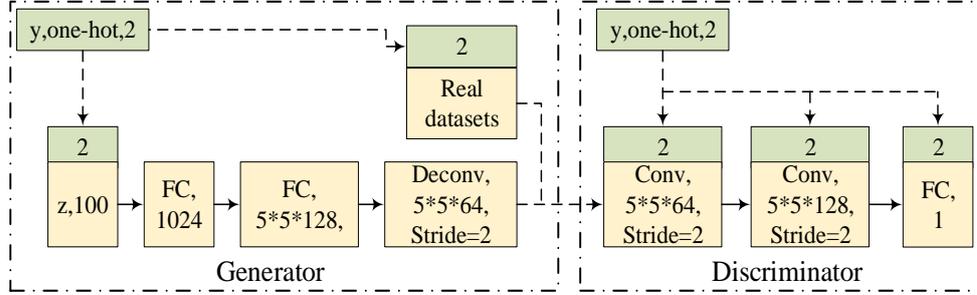

Fig. 5.   The architecture of LSGAN

Based on the architecture of LSGAN, how data augmentation happens will be described clearly as follow. The class labels obtained by SFCM are converted into 2-dimensional vector by using one-hot encoding. The 2-dimensional vector combined with the 100-dim noise vector $z$ is used as the input of the generator. As shown in Fig. 5, the generator of the LSGAN is made up of 2 FC layers and 1 deconvolution layer. Firstly, the 102-dimensional input vector is reshaped into 49*2*64 dimensions through 2 FC layers. And then, it is fed into the deconvolution layer with the kernel size of 5*5*64 (height 5, width 5, number of channels 64, and stride 2). Finally, the output of deconvolution layer is a 3-dimensional feature tensor with size 97*3*1, which is the same size for the generated samples.

For the discriminator of LSGAN, either real or generated data is used as its input. The discriminator consists of 2 convolutional layers and 1 FC layer. The kernel size of 2 convolutional layers are respectively 5*5*64 and 5*5*128. The 2 convolutional layers map the input into a 24*1*128 tensor. Finally, the last FC layer outputs a 1-dimensional vector which represents the probability of the input information stemming from the real data.

In this way, the generator and discriminator are trained through an alternating iteration procedure and finally reach a Nash equilibrium, which can generate fake samples as real as possible for data augmentation. Thus, at the offline training stage, the BiGRU-Attention-based STVSA model can fully capture the time dependencies on the basis of the augmented big training dataset.

*C. Offline Training*

The architecture of the BiGRU-Attention based assessment model at the offline training stage is shown in Fig. 6.

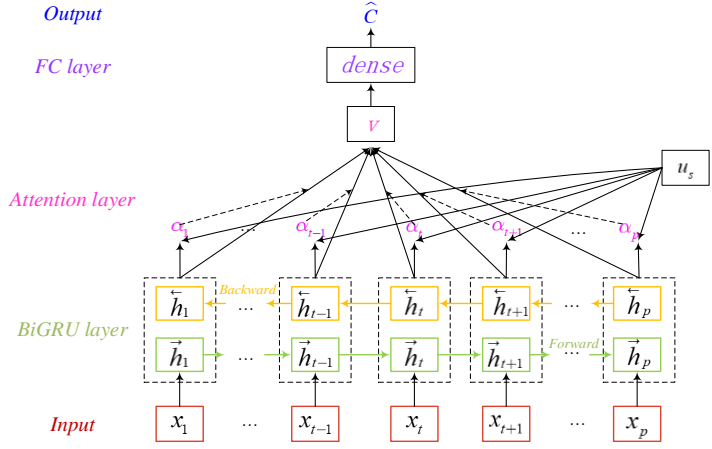
Fig. 6. BiGRU-Attention model

The BiGRU-Attention model used in this paper mainly includes 5 layers, which are the input layer, BiGRU layer, attention layer, FC layer, and output layer.

For each sample, the input of the BiGRU-Attention model is all buses' TSs constituted by input features and at each sampling moment, it is an $m$-dimensional feature vector $x_t = \{U_{t,1}, \cdots U_{t,L}, P_{t,1} \cdots P_{t,L}, Q_{t,1} \cdots Q_{t,L}\}$.

The BiGRU layer aims to extract the latent temporal dependency from the input features. Different from an original GRU, the hidden state $h_t$ is influenced by both the forward hidden state $\vec{h}_t$ and the backward hidden state $\overleftarrow{h}_t$ in the BiGRU. In this way, the BiGRU is capable of more fully capturing the dependency.

The input of the attention layer is the activated output vector $h_t$ of the BiGRU layer. In the attention layer, the strong correlated hidden features will be automatically allocated high weights and vice versa, which improves the assessment performance through emphasizing the significant information which affects the stability status.

The FC layer in the BiGRU-Attention is used to output the stability results, which is shown as
$$C = \text{softmax}(W_{dense} \cdot s + b_{dense}) \quad (20)$$
where $s$ denotes the input of the FC layer; $W_{dense}$ is the weight matrix; $b_{dense}$ represents the bias matrix; the softmax() refers to the activation function. The output of the FC layer is the final STVSA result of the built BiGRU-Attention model.

For a given dataset $\{X(j), y(j)\}_{j=1}^{N}$ with $N$ training samples, the offline training based on the BiGRU-Attention model aims to obtain network parameters $W$, $U$, and $b$. The Adam optimizer is adopted to find the optimal value of these parameters, and the Euclidean distance with $L_2$ norm is used as the loss function.

### D. Online Application

At the online application stage, real-time measurements are collected by phase measurement units (PMUs). Once the measurements are acquired, it will be fed into the assessment model which obtains its optimal parameters through offline training. And then, the STVS result of a system can be determined immediately. If the assessment result indicates that the system isn't able to maintain stability status, remedial control measures must be taken to prevent the system from voltage collapse at once; otherwise, the assessment model continues to monitor the system stability status. Note that at the online application stage, the trained model can be updated periodically to improve the adaptability of the proposed intelligent system for unexpected cases under various operating conditions.

### E. Evaluation Indicators

To properly evaluate the quality of the generated samples obtained by the LSGAN-based data augmentation and the performance of the BiGRU-Attention based assessment model, the following evaluation indicators are introduced in this work.

1) Evaluation indicators of data augmentation

In this paper, the Wasserstein distance (WD), Maximum mean discrepancy (MMD), and Fréchet inception distance (FID) are adopted as the quantitative evaluation indicators to validate the diversity and quality of the generated samples obtained by the LSGAN based data augmentation.

a) WD

The WD is used to measure the affinity between real data distribution $P_r$ and generated data distribution $P_g$, which is defined as [36]:

$$\text{WD}(P_r, P_g) = \inf_{\gamma \sim \prod(P_r, P_g)} E_{(x^r, x^g) \sim \gamma} [d(x^r, x^g)] \quad (21)$$

Here, $\prod(P_r, P_g)$ is the set of joint distributions, in which the marginal are respectively $P_r$ and $P_g$. A low WD indicates that the 2 distributions are similar, and the quality of generated samples is good.

b) MMD

The MMD measures the dissimilarity between the 2 probability distributions, which is widely applied to evaluate the quality of the generated samples. The MMD of 2 data distributions $P_r$ and $P_g$ is given by [36]:

$$\text{MMD}(P_r, P_g) = E_{(x_r, x_r') \sim P_r}[k(x_r, x_r')] - 2E_{\substack{x_r \sim P_r \\ x_g \sim P_g}}[k(x_r, x_g)] + E_{(x_g, x_g') \sim P_g}[k(x_g, x_g')] \quad (22)$$

In (22), $k$ is a fixed kernel function. If the 2 distributions $P_r$ and $P_g$ are closer, the value of MMD is lower, indicating a better quality of generated samples.

c) FID

Based on the convolution feature of the Inception network, the FID models the real data distribution $P_r$ and the generated data distribution $P_g$ as a Gaussian distribution with means $\mu_r$ and $\mu_g$ and empirical covariances $\text{cov}_r$ and $\text{cov}_g$. The FID is computed as follows [37]:

$$\text{FID}(P_r, P_g) = \|\mu_r - \mu_g\| + \text{Tr}[\text{cov}_r + \text{cov}_g - 2(\text{cov}_r \text{cov}_g)^{1/2}] \quad (23)$$

Note that the lower the value of FID, the better the quality and diversity of generated data. Meanwhile, the FID can also effectively reflect the problem of model collapse.

2) Evaluation indicators of STVSA model

Besides accuracy, this paper also uses other statistical indicators such as AUC, MCC, and F1-score to evaluate the performance of the STVSA model.

a) Accuracy

Accuracy is a commonly-used index to evaluate the performance of the STVSA model [4, 5, 12-15].

$$\text{Accuracy} = \frac{TP + TN}{TP + FP + FN + TN} \quad (24)$$

In (24), the accuracy is the proportion of the correctly predicted samples by the model, which evaluates the performance of the proposed method as a whole.

b) AUC

AUC is the area under the receiver operating characteristic (ROC) curve which is a 2-dimensional depiction of classifier performance [38]. And the larger the value of AUC, the better the performance of the classification model. The horizontal and vertical axes of the ROC curve are as follows [22, 29, 38]:

$$\text{TPR} = \frac{TP}{TP + FN} \quad (25)$$

$$\text{FPR} = \frac{FP}{FP + TN} \quad (26)$$

where TPR denotes the proportion of all stable samples that are correctly predicted to be stable; FPR is the proportion of all unstable samples that are misjudged as stable.

c) MCC

MCC is considered to be an important statistical indicator for evaluating the classification performance of stable/unstable samples of different methods, which is defined as [30]:

$$\text{MCC} = \frac{TP \times TN - FP \times FN}{\sqrt{(TP + FP)(TP + FN)(TN + FP)(TN + FN)}} \quad (27)$$

The MCC value is between -1 and +1, the value of +1 means perfect prediction, and -1 means that the prediction result is completely inconsistent with the real situation.

d) F1-score

The F1-score is a weighted average of the precision and recall for a specified confidence threshold to estimate overall classification accuracy, which is given as follows [22, 30, 38].

$$precision = \frac{TP}{TP + FP} \quad (28)$$

$$recall = \text{TPR} = \frac{TP}{TP + FN} \quad (29)$$

$$\text{F1-score} = 2 \cdot \frac{precision \cdot recall}{precision + recall} \tag{30}$$

## V. Case Study

To verify the effectiveness and superiority of the proposed method, it has been examined through a series of comprehensive tests on the New England 39-bus system, which is widely used in the field of STVSA [12, 14, 15, 22]. The main hyper-parameters of LSGAN and BiGRU-Attention are respectively given in Tables II and III. Note that the parameters are chosen by the try-and-error method in this study.

TABLE II
Hyper-paraments setting of LSGAN

| Hyper-parameters | Values |
|---|---|
| learning rate | 0.0001 |
| beta1 | 0.5 |
| k | 4 |
| batch size | 32 |
| epoch | 100 |
| iteration | 3000 |

In Table II, the learning rate and beta1 of the Adam optimizer are respectively set to $1 \times 10^{-4}$ and 0.5; Generally speaking, discriminator needs to be trained more than the generator, $k$ is the parameter used to control the balance of discriminator and generator; 32 batch sizes were used to maximize the usage of GPU resources. The total number of training epochs is 200; the number of iterations is 3000.

TABLE III
Hyper-paraments setting of BiGRU-Attention

| Hyper-parameters | Values |
|---|---|
| learning rate | 0.0001 |
| dropout | 0.25 |
| hidden layer unit | 64 |
| batch size | 64 |
| epoch | 200 |
| attention size | 8 |

From Table III, it can be found that the values of learning rate in LSGAN and BiGRU-Attention are the same; the number of hidden layer units of BiGRU-Attention network is 64; the value of batch sizes is 64; to prevent overfitting, a dropout of 25% for the BiGRU and FC layer is implemented [39]. The BiGRU-Attention based model is trained for 200 epochs. Attention size represents the linear size of the attention weights.

All experiments have been implemented via the Google TensorFlow 1.14.0 on a windows platform with an Intel Xeon CPU E5-2678 v3, an NVIDIA GeForce RTX 2080 Ti GPU and 64GB RAM. The programming language used in this study is Python 3.7. Note that PMU data is simulated through detailed TDSs by using the commercial power system simulation software PSD-BPA.

### A. Generation of Original Small Dataset

In this study, detailed TDSs are employed for generating the original small dataset. The STVS status of the system is strongly correlated with the dynamic characteristics of fast acting load components like an induction motor and an exciter after a large disturbance [1]. To this regard, a typical induction motor is considered in the dynamic modeling of loads. To cover different contingencies and operating conditions, a variety of different operating conditions, including fault location, fault type, fault clearing time, and proportions of dynamic load, are considered in the TDSs.

(i). The total load demand is respectively set to 80%, 100%, and 120% of the base level;

(ii). All loads employ a composite load model consisting of static ZIP loads and motor loads in parallel, where the proportions of motor loads are respectively set to 60%, 70%, 80%, and 90%;

(iii). A 3-phase short-circuit fault is imposed on each transmission line as the fault type adopted in this study.

(iv). The short-circuit faults are respectively set to located at 0%, 20%, 40%, 60%, 80% of a certain transmission line;

(v). A fault clearing time is varied in the range [0.15 s, 0.2 s] when a short-circuit fault occurs at 0.1 s.

Based on the above setting, the original small dataset including 1200 samples is got by the TDSs, and the sampling time is 0.01s. By adding the generated cases stemming from data augmentation into the original small dataset, the final dataset with total 10640 samples is obtained for the subsequent analysis. In this study, the original small dataset is divided into the training dataset and the testing dataset randomly according to a ratio of 3:1, which accords with [4, 12, 22]; so does those in the final dataset.

## B. Performance Test of SFCM

To examine the effectiveness of the SFCM, it is compared with the commonly used constraint-partitioning k-means (COP-k-means) [4]. According to the silhouette coefficient, the computation results of these semi-supervised cluster learning algorithms are shown in Table IV.

TABLE IV
Comparison of different clustering algorithms

| Index | SFCM | COP k-means |
|---|---|---|
| silhouette coefficient | **0.3856** | 0.3482 |

It can be seen from Table IV that the SC value of the SFCM is higher than that of the COP-k-means, which suggests that hidden rules contained in datasets can be better mined, and a set of more reliable class labels can be obtained by the SFCM.

## C. Performance Test of LSGAN

In this part, WD, MMD, and FID are employed to examine the performance of the proposed deep adversarial data augmentation in terms of affinity and diversity. Besides, to verify the validity of LSGAN used in this paper, we compared it with the conditional GAN (CGAN) under the same condition. Furthermore, to conduct a reliable assessment, all experiments are performed 3 times with different random seeds, and the means are reported in Table V.

TABLE V
WD, MMD and FID comparison of LSGAN and GAN

| Evaluation indicators | LSGAN | CGAN |
|---|---|---|
| WD | 4.56 | 5.31 |
| MMD | 0.06 | 0.062 |
| FID | 2.19 | 2.40 |

Compared to the CGAN, it can be seen that LSGAN model improves the WD value by 21.9%, MMD value by 3.23%, and FID value by 8.75% respectively, showing the superior performance in terms of affinity, quality, and diversity. In other words, LSGAN model can generate the synthetic data with high-quality data, demonstrating its outstanding advantages of data augmentation. And then, the accuracies of the presented DLBAN before and after data augmentation are shown in Table VI.

TABLE VI
Accuracy comparison of before and after data augmentation

| Dataset | Accuracy (Final dataset) | Accuracy (Original dataset) |
|---|---|---|
| Testing dataset | 99.44% | 95.76% |
| Training dataset | 99.51% | 95.94% |

From Table VI, it can be found that the training and testing accuracies have been significantly improved via LSGAN-based data augmentation. This fact suggests that data augmentation is an effective tool that enables the proposed deep learning-based assessment model to work well with small datasets.

## D. Performance Test of BiGRU-Attention Based Model

To reasonably evaluate the performances of the proposed model, comparative tests with other deep learning networks (LSTM and GRU) and shallow machine learning networks (DT and Support Vector Machine (SVM)) by using evaluation indicators including accuracy, AUC, MCC, and F1-score. Here, the parameter settings of LSTM and GRU are the same as that of BiGRU-Attention, while the SVM adopts the grid search together with cross-validation to determine its optimal parameters. The test results of different algorithms are shown in Table VII.

TABLE VII
Comparison about accuracy of different approaches

| Approach | Accuracy |
|---|---|
| **BiGRU-Attention** | **99.44%** |
| GRU | 97.14% |
| LSTM | 96.39% |
| DT | 92.33% |
| SVM | 84.33% |

It can be seen from Table VII that the accuracy of the BiGRU-Attention is the highest, and that the accuracies of the GRU and LSTM are also higher than those of the DT and SVM. This demonstrates that learning from temporal dependencies of the quantified TSs contributes to obtain high accuracies. Furthermore, the BiGRU-Attention outperforms the GRU and LSTM via the bidirectional information transmission and attention mechanism in this study.

To further comprehensively measure the superiority of the proposed method, the performance of the proposed method is further analyzed and discussed based on some statistical indicators, as shown in Table VIII.

TABLE VIII
Comparison about MCC, F1-score and AUC of different approaches

| Approach | MCC | F1-score | AUC |
|---|---|---|---|
| **BiGRU-Attention** | **0.9888** | **0.9945** | **0.9993** |
| GRU | 0.9432 | 0.9653 | 0.9894 |
| LSTM | 0.9285 | 0.9539 | 0.9879 |
| DT | 0.8406 | 0.9363 | 0.9510 |
| SVM | 0.6669 | 0.8773 | 0.9356 |

As far as the MCC is concerned, its value always exceeds 0.98 by using deep learning, where the MCC of the BiGRU-Attention is 4.83% higher than the GRU, 6.49% higher than the LSTM. In contrast to the DT and SVM, it is respectively increased by 17.63% and 48.27%. And so do this in the test results in terms of the F1-score and AUC, i.e., the indicator value of the BiGRU-Attention is not only much higher than those of the DT and SVM but also much better than those of the GRU and LSTM. These results further validate that the proposed method outperforms the widely used deep learning algorithms (GRU, LSTM) and traditional shallow learning ones (DT, SVM).

Besides accuracy, the costs of misdetections and false alarms are another concerns when using deep learning for STVSA. The misclassification costs of different assessment models are shown in Table IX.

TABLE IX
Comparison about accuracy Misclassification costs of different approaches

| Approach | Condition | Prediction accuracy on the testing dataset | |
|---|---|---|---|
| | | Classified as stable(%) | Classified as unstable(%) |
| BiGRU-Attention | Stable | 98.92%(1368/1383) | 1.08%(15/1383) |
| | Unstable | **0%(0/1277)** | 100%(1277/1277) |
| GRU | Stable | 95.88%(1326/1383) | 4.12%(57/1383) |
| | Unstable | 1.49%(19/1277) | 98.51%(1258/1277) |
| LSTM | Stable | 94.65%(1309/1383) | 5.35%(74/1383) |
| | Unstable | 1.72%(22/1277) | 98.28%(1255/1277) |
| DT | Stable | 92.35%(169/183) | 7.65%(14/183) |
| | Unstable | 7.69%(9/117) | 92.31%(108/117) |
| SVM | Stable | 91.80%(168/183) | 8.20%(15/183) |
| | Unstable | 27.35%(32/117) | 72.65%(85/117) |

Taking the costs of misdetections and false alarms into account, the reliability of the presented approach is verified in Table IX. In particular, the misdetection rate of the proposed approach is 0%, which is superior to those of the others. For STVSA, a high false alarm rate will cause some economic losses but will not impair the STVS of the system; while a high misdetection rate will make the system take no measure to prevent possible instability accidents, which may lead to disastrous consequences. Hence, a conclusion may be safely drawn that the 0 misdetection rate and acceptable false alarm rate obtained by the proposed model confirm its reliability.

*E. Computational Efficiency Analysis*

Taking into account that the STVS issue is a fast phenomenon of the order of several seconds [1], it is critical to detect stability results quickly. Since the online assessment time of the trained Furthermore, considering that the reliable remedial control needs to be started up within a short time, a proper OTW is of an essential impact on the performance of the classification. A smaller window size will lead to a rapid, but inaccurate assessment; while a larger size window will lead to an accurate, but untimely assessment. By performing a sensitivity analysis, the performance of the BiGRU-Attention model is tested under the OTWs with different lengths, as shown in Fig. 7.

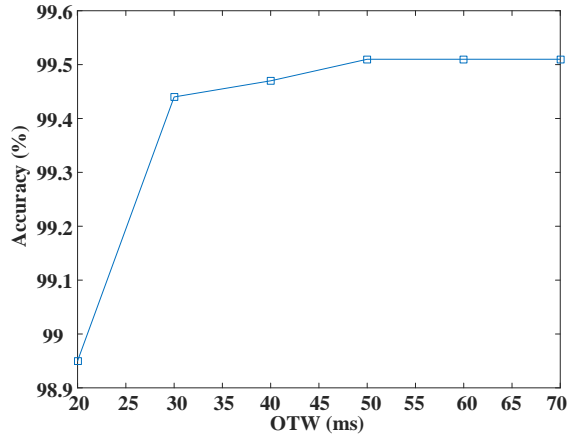

Fig. 7. Accuracies under different OTWs

From Fig. 7, it can be observed that when the length of the OTW exceeds 30 ms, the accuracy of the BiGRU-Attention-based STVSA model only marginally rises. Therefore, to achieve the trade-off between assessment accuracy and rapidness, the OTW length is selected as 30 ms, which outperforms other existing state-of-the-art methods with a shorter response time and no sacrifice of accuracy [5, 13-15]. Besides, the BiGRU-Attention model only needs 0.069 ms to obtain the assessment results during online monitoring, which is a desirable speed. And compare with the length of the OTW for data acquisition, it can be ignored.

According to the IEEE Standard C37.118.2-2011, the typical communication delay between PMU and phasor data concentrator (PDC) is 0.02 to 0.05 second [40]. Consequently, taking the communication into account, the maximum response time is no more than 0.08 second, which reserves sufficient time for the measurement, electrical communication, and other delays from PDCs to a dispatch center of power systems. Furthermore, it can ensure the immediate decision-making and subsequent control actions against the short term voltage instability.

*F. Robustness Test under Noisy Environments*

Since noises are inevitable during the collection and transmission process of PMU data [20], the robustness of the proposed method is tested under noisy environments. Here, Gaussian white noises with different signal-to-noise ratios (SNRs) are added into the PMU data, where a smaller SNR indicates a higher noise level. Here, 3 scenarios with SNRs 50 dB, 40 dB, and 30 dB are respectively tested in this section. In terms of accuracy, AUC, MCC, and F1-score, the test results under noisy environments are listed in Table X.

TABLE X
Performance comparison under different SNRs

| SNR | Accuracy | MCC | F1-score | AUC |
|---|---|---|---|---|
| 50dB | 99.40% | 0.9880 | 0.9944 | 0.9976 |
| 40dB | 99.36% | 0.9872 | 0.9939 | 0.9948 |
| 30dB | 99.32% | 0.9865 | 0.9939 | 0.9905 |

It can be seen from Table X that although the assessment accuracy will decrease slightly with the increase of the background noise level, it is always above 99% in various noise environments. Similarly, the other 3 indicators MCC, F1-score, and AUC also hold a very high level, which fully verifies the excellent robustness of the proposal against noises.

# VI. Conclusion

Deep learning has been proven as a promising technique for STVSA. As current state-of-the-art deep learning algorithms suffer from the inapplicability of a small training dataset with insufficient data due to expensive and trivial data collection and annotation, how to make a deep learning-based STVSA model work well on a small training dataset is a challenging and urgent problem. Although a big enough dataset can be directly generated by contingency simulation, data augmentation is a superior and irreplaceable proposition by artificially inflating the representative and diversified training datasets in low-cost and efficient way. In view of this, this paper proposes a deep-learning intelligent system incorporating data augmentation for the STVSA and performs simulations on the IEEE 39-bus system. Based on the test results, the main conclusions of this paper are as follows:

(1) By leveraging data augmentation, the proposed deep learning framework DLBAN enables a deep learning-based assessment model to work well with small training datasets. This provides a general tool for enabling deep learning to handle the problem of the unavailability of big data.

(2) The BiGRU-attention-based assessment model is able to not only fully capture the temporal dependence in time series data, but also improve the feature extraction ability via attention mechanism.

(3) The test results demonstrate that the proposed approach outperforms other state-of-the-art alternatives with better accuracy and faster response time. Through implementing statistical tests, the superiority of the DLBAN has been further validated by using statistical indicators AUC, MCC, and F1-score besides accuracy.

(4) For a small-scale dataset, LSGAN-based data augmentation manages to significantly improve the training and testing accuracies in our study. Besides, the test results under noisy environments verify the good robustness of our proposed approach.

In future work, the problem of missing PMU information will be considered for STVSA. Besides, efforts will be devoted to eliminating the impact of the class imbalance problem caused by rare instability events. Another interesting topic is to use automated reinforcement learning (Auto-RL) to maximize the predictive performance of the presented STVSA model by automatically determining the most suitable model architecture and hyperparameters [41].

**Acknowledgements**

This work is partly supported by the Natural Science Foundation of Jilin Province, China under Grant No. YDZJ202101ZYTS149.